\documentclass[a4paper, 10pt, conference]{ieeeconf}      % Use this line for a4
\IEEEoverridecommandlockouts                              % This command is only
\usepackage[utf8]{inputenc}
\usepackage[T1]{fontenc}
\usepackage{changes}
\usepackage{todonotes}
\title{\LARGE \bf
Rethinking Trust in Social Robotics
}

\author{%
Rachele Carli$^{1}$ and Amro Najjar$^{2}$% <-this % stops a space
\thanks{$^{1}$R. Carli is with the Research Institute for Human-Centered AI, University of Bologna, Bologna, Italy
        {\tt\small rachele.carli2@unibo.it}}%
\thanks{$^{2}$A. Najjar is with the AI Robolab/DCS, at the University of Luxembourg, Esch-sur-Alzette, Luxembourg
        {\tt\small amro.najjar@uni.lu}}%

}
\begin{document}

\maketitle
\thispagestyle{empty}
\pagestyle{empty}

%%%%%%%%%%%%%%%%%%%%%%%%%%%%%%%%%%%%%%%%%%%%%%%%%%%%%%%%%%%%%%%%%%%%%%%%%%%%%%%%
\begin{abstract}

In 2018 the European Commission highlighted the demand of a human-centered approach to AI. Such a claim is gaining even more relevance considering technologies specifically designed to directly interact and physically collaborate with human users in the real world. This is notably the case of social robots. The domain of Human-Robot Interaction (HRI) emerged to investigate these issues."Human-robot trust" has been highlighted as one of the most challenging and intriguing factors influencing HRI. On the one hand, user studies and technical experts underline how trust is a key element to facilitate users' acceptance, consequently increasing the chances to pursue the given task. On the other hand, such a phenomenon raises also ethical and philosophical concerns leading scholars in these domains to argue that humans should not trust robots.

However, trust in HRI is not an index of fragility, it is rooted in anthropomorphism, and it is a natural characteristic of every human being. Thus, instead of focusing solely on how to inspire user trust in social robots, this paper argues that what should be investigated is to what extent and for which purpose it is suitable to trust robots. Such an endeavour requires an interdisciplinary approach taking into account (i) technical needs and (ii) psychological implications.

\end{abstract}

%%%%%%%%%%%%%%%%%%%%%%%%%%%%%%%%%%%%%%%%%%%%%%%%%%%%%%%%%%%%%%%%%%%%%%%%%%%%%%%%
\section{INTRODUCTION}
Nowadays, social robotics is becoming increasingly pervasive. 
Machines implemented with artificial intelligence (AI) are no longer exclusive prerogative of engineers or industrial use. Social robots are increasingly part of our everyday reality, able to act in an unstructured environment, directly interacting with inexperienced users. \cite{pandey2016socially}. Some of them are designed to be used in direct contact with fragile individuals – like the elderly \cite{wu2012designing}, children \cite{toh2016review} or disabled \cite{hersh2015overcoming}–, some others to work together with human teammates \cite{groom2007can}, even in high-risk environments \cite{voth2004new}. 
This raised the need on one hand to equip them with social competences as accurate as possible, on the other to identify the elements that could favour users’ acceptability \cite{gaudiello2016trust}. 

Trust is one of the basis for  any functional relationship and for this very reason it became an element of major interest in the field of social robotics. In fact, it was demonstrated that trusting the machine facilitates interaction, allowing it to perform the given task more efficiently and making individuals more comfortable \cite{freedy2007measurement}. 
Trust measures fall into explicit measures - involving physical appearance and social cues, as gazing, gesture, sound of voice – and implicit ones - including impressions, beliefs, preconceptions of the individual involved - \cite{yagoda2012you}.
In line with this assumption, it was noted that people are more likely to accept and efficiently collaborate with robots that have characteristics more akin to human ones, instead of merely referable to machines \cite{toure2015or}. 
For this reason some robots are designed to simulate a doubtful and uncertain behaviour or to hesitate before performing a task, for these attitudes are typical of humans rather than artificial objects. An example could be the recent research related to the “inner speech functionality” implemented in robots, which allows them to emulate the human act of "talking to themselves" in order to focus or to evaluate how to behave \cite{pipitone2021robots}. It was noticed that this makes the devices appear more competent in their problem solving activities and to appear more trustworthy  from the human-user standpoint, since they are demonstrating a human-like behaviour \cite{pipitone2021robots}. In fact, the inner-voice is usually associated with the possess of self-consciousness and self-awareness, prerogative of the humankind \cite{morin2009inner}.
On closer inspection, such dynamics prove effective from a functional point of view, but they can also have ethical repercussions that should not be underestimated, connected to the themes of deception, manipulation, dehumanisation and over-trusting dynamics \cite{wilson2016reflections}.
In light of this, a part of the philosophical debate is directed at the idea according to which people “should not trust AI systems” \cite{ryan2020ai}. However, from a multidisciplinary perspective, such a statement could sound reductive for it is important, indeed, to consider also factual aspects. Looking at concrete HRI dynamics, it is evident that human beings are naturally led to develop relationships of intimate trust with what they interact with repeatedly over time.
This analysis suggests another perspective through which to approach the “human-robot trust” field. To this end, it takes into account both the needs of technicians and the respect for the human person, understood in the totality of their values and rights.

As this paper aims to highlight, one solution could be to consider not only whether or just how, but to what extent it is suitable for the users to trust social robots. 

To this end, (section II) analyses the concept of trust, to underline its non-univocality and its bond with the psychological functioning of human beings. Then it contextualises this very notion in the HRI domain. (Section III) takes into account the claim according to which people should not trust robots, from a philosophical perspective. However, (section IV) considers whether it is really possible to prevent or deny the tendency individuals have to confide in machines. For this purpose the phenomenon of anthropomorphism is introduced. Then, (section V) highlights the controversial differentiation between the concept of trust and the one of trustworthiness, in order to outline the respective areas of operation. This will be functional to (section VI), which suggests an alternative to the dominant "no trust-no use" approach to social robotics, through the here proposed change in prospective. In (section VII) the fruitful contribution this proposal is discussed, in order to underline its role in framing future researches.

\section{TRUST FROM A MULTIDISCIPLINARY PERSPECTIVE}
Trust is a very debated concept, for it does not have a univocal understanding among scholars. It has many definitions, which vary depending on the domain that is taken into account. Hence, its boundaries are blurred and often overlapped on other notions, similar but not corresponding, such as trustworthiness and reliance.
What is clear is that trust is a typical strategy that humans use to cope with uncertainty \cite{botsman2017can} and it has some roots in what is known – past experiences, regarding both the trustee and the trustor themselves – and in what can be predicted – future, not unexpected events –. 
Therefore, the idea of trust as the result of a mere cognitive-based process - focused on the final goals and a “gains-costs” balance - is not \textit{per se} sufficient \cite{taddeo2010modelling}. In fact, it has been proved that some categories of individuals, like parents or relatives, are statistically trusted more than non-family members, on equal objective terms or even postponing the evaluation of their effective worthiness \cite{white2005impact}. On this basis, it can be assumed that any evaluation regarding previous performances which creates reasonable expectation is a matter of reliance only \cite{tuomela2003simulating}.
On the contrary, trust is grounded on confidence about another’s good will and benevolence towards the trustor, based on affective attitude \cite{jones1996trust}. Thus, in such a view, the free will of the trustee is fundamental and brings with itself the possibility to be betrayed. This implies that the one we trust must have the option to act unlike how we would expect or even not to act at all. 
In this case, the trustee must be aware of both the expectations the trustor has towards him and what should be done in order not to disregard them \cite{ryan2020ai} . It follows that the ones trusted have to be blameable for the breach of trust or – said otherwise – they have to be considerable responsible for their actions \cite{bryson2018ai}
Based on that, it is important to investigate whether this conception of trust - as delineated for the human-human realm - can be extended even to the HRI context.

\subsection{Trust in Artificial Intelligence}
The need of some sort of control mechanisms in artificial societies has been discussed since the early days of artificial intelligence. Traditionally, computational security has been used to deal with set of well-defined threats, by relying on cryptography algorithms \cite{forouzan2011cryptography}. Yet, this approach is highly centralised and can be brittle. To overcome this problem, several other "soft control" techniques have been proposed. Namely, social control mechanisms (e.g., trust and reputation) has been outlined by the multi-agent systems (MAS) community as efficient mechanisms that do not prevent undesirable events, but ensure some social order in the system, without restricting the system development. Thus, by enforcing these mechanisms, they enable the system to evolve while preventing these anomalies from occurring again \cite{castelfranchi2000engineering}. The definition of trust proposed by Gambetta et al. \cite{gambetta2000can} is among the most widely used definitions of trust in the AI and MAS community: "Trust is the subjective probability by which an agent A expects that another agent B performs a given action on which its welfare depends". 

Based on this definition, the use of trust in MAS and AI is twofold. First, it is a mental assessment on how trustworthy other agents are, and second, it is an intention or a decision, based on that assessment, to trust that agent or not e.g., delegate a critical task to it \cite{calvaresi2019explainable, sabater2013trust}.  

\subsection{Trust in HRI}
The above explained conceptualisation is valid for artificial intelligence in general and for MAS. However, when it comes to social robotics we need to take into account a new dimension of trust, for this field implies an intimate interaction between the machine and the user. 
In fact, these devices are designed to progressively enter our everyday life, being our companions, caregivers, entertainers. However, researchers still lack a precise and fully predictable understanding of the mechanisms behind their performances and the very nature of the outcome produced, posed a given input. This have led to the proliferation of investigations aiming to enhance trust in technology, from a "no trust-no use" perspective. 
Among the most detailed and broadly accepted definitions of trust in the HRI sector we can find: the one which identifies it as “the attitude that an agent will help achieve an individual’s goals in a situation characterised by uncertainty and vulnerability” \cite{lee2004trust} and the one which describes it as “a belief, held by the trustor, that the trustee will act in a manner that mitigates the trustor’s risk in a situation in which the trustor has put its outcomes at risk” \cite{wagner2011recognizing}.

It follows that the attention of the analysis of such a dynamic is on the trustor, since it is the one to whom the attitude belongs and who can address it to the machine or not. However, the users of social robots are common, non-expert people, with different age, cognitive levels, cultural backgrounds. 
This makes them possibly vulnerable to episodes of misrepresentation and manipulation, that rise concerns in the ethical and psychological domains. 
These will be analysed in (section III) and (section IV), so as to give a more complete understanding of the "human-robot trust" phenomenon. 

\section{PHILOSOPHICAL CONCERNS: SHOULD WE TRUST ROBOTS?}
The main interpretations of the notion of trust in the literature highlighted the relevance of the trustee's (i) emotional disposition, (ii) freedom of choice and (iii) faculty to understand and be held responsible for the consequences of their choices. Accordingly, some recent works in humanities argue that it is not possible to reconcile this concept with dynamics affecting the interaction between humans and artefacts \cite{coeckelbergh2012can}.
Despite their pleasant design, their social competences and different degrees of – weak – autonomy \cite{gutman2012action}, social robots are nothing more than very sophisticated, innovative objects \cite{bertolini2013robots}.
It is still subject of debate to what extent these robots are able to perceive their own existence, to settle autonomously a goal and to experiments either goodwill or maleficence, for they cannot have free will. 
It could be argued that such a statement is affected by the opacity often related to the black box phenomenon \cite{szegedy2014intriguing}, which would make difficult, or sometimes impossible, for the programmers to foresee and have full control over the outcome of the machine, given a precise input \cite{villani2018meaningful}. 
This has led to the so called “no trust-no use” approach to social robotics. According to it, trusting machines is the main way to foster their very adoption. In fact, it would encourage the user’s participation and so facilitate the artefact in fulfilling its task.
However, it has to be underlined that, notwithstanding the final result and the external appearance, robots’ behaviour is always the result of the way they were programmed. Therefore, there is always a human being to be considered liable for the outcome produced by an AI system \cite{gunkel2012machine}.  

Consequently, with regards to an artefact, we should refer to a form of “translated trust”, instead of a proper one. 
In this view, trust in technology is nothing more than a consequence of trusting the expert behind it \cite{pitt2010s}. In fact, what people expect from a machine is that it works correctly, is suitable for the purpose for which it was built, and is safe in its use. Put another way, human beings ask the artificial to remove the uncertainty inherent in their fickle nature, to be efficient and reliable in a way that is alien to humankind. They do not ask it to be a faithful reproduction of themselves and their interpersonal dynamics \cite{bisol2014ethics}.

Nonetheless, radically stating that people should not trust social robots means to underestimate that trust is ultimately an emotional response and that people feel it, before cognitively represent it. In order to have a comprehensive understanding of this phenomenon it is appropriate to deepen the subconscious mechanisms that regulate the human mind, influencing the interaction with robots. 

\section{PSYCHOLOGY OF TRUST AND ANTHROPOMORPHISM: CAN WE TRUST ROBOTS?}
People are born with the tendency to trust and distrust. This implies that our personalities and past experiences can influence our attitude to the act of trusting. Nevertheless, the main role is played by the trustees – by what they do, but especially what they evocate to the trustor – \cite{botsman2017can}. In general, it was demonstrated that we usually trust people similar to ourselves, events recalling previous events, and people or systems that we have already, successfully, trusted \cite{gerrig2008psychologie}.
 
For what social robotics is concerned, the search for similarity is widely filled through anthropomorphism. It consists in the natural propensity people have to attribute human-like features to the inanimate objects they interact with repeatedly over the time \cite{hegel2008understanding}. With regards to social robots, such elements are not merely attributed – as in the case of toys – but inferred, on the base of specific design choices \cite{damiano2018anthropomorphism}. Despite previous theorisations of anthropomorphism as a bias – common among fragile or still cognitively immature subjects – this phenomenon is now valuated as an essential, inherent component of human-mind \cite{levillain2017behavioral}. 

Nevertheless, even considering it as a weakness, we should heed that vulnerability was described as the core of humanity. It cannot be radically eliminated, for it is reflected in many aspects of our psyche and nature \cite{coeckelbergh2013human}. One of them is the innate propensity – regardless of personal individuality – to trust itself. Part of contemporary psychology interprets this as an unconscious need to feel vulnerable to others, to rely on others, not to be entirely self-sufficient. Nonetheless, Aristotle had already defined humans as "social animals", driven to aggregation. Freud had then introduced the category of humans as "symbolic animals", for led to conceptualise material reality through symbolic interpretation \cite{cassirer1996filosofia}. 
It follows that we cannot get rid of anthropomorphism and, together with it, of the disposition to create with robot bonds that should be more appropriate among people only. 

\section{TRUST \textit{versus} TRUSTWORTHINESS}
As it was previously analysed (Section I), trust consists in confiding in someone to carry out a specific action. It is surely influenced by circumstances \cite{cvetkovich2002new} but it is, ultimately, a personal state, a subconscious propensity \cite{mayer1995integrative} that transcends any factual element and has its roots in the model of attachment developed during childhood \cite{schoorman2007integrative}.
It follows that it could be fruitful not to focus on trust - only - but on a different criterion: trustworthiness. 
Being trustworthy can convey a trust bond, but it is not \textit{per se} sufficient. While trust is an attitude which depends also on personal and emotional elements,  trustworthiness is a property of the object itself. It is connected to competence, efficiency and so with objective and valuable characteristics \cite{castelfranchi2010trust} . 
Therefore, if trust is something that must be earned, trustworthiness is a characteristic that can be projected and materially implemented from the outside. For this reason, if the first one is a dominant element of our interpersonal relations, the second one should be the central component of our interaction with social robots. 
Such a distinction risks to be seen as a mere speculative exercise. Otherwise, setting correctly the differences among these two concepts and defining their spheres of expertise are the starting points for any technological analysis that aspires to lay the foundations for an effective and pervasive development.  

\textbf{Trustworthiness in HRI.}
The HRI literature shows us that a user may not trust a robot that reflects the technical standards of reliability and \textit{vice versa}. The same robot can be trusted by others, just because - external factors being equal - their emotional and psychological past experiences are different \cite{wagner2018overtrust}

On the contrary, if trustworthiness is a property, it can be implemented in technological devices by design. This means that a specific machine will be considered suitable - or not - for the task it has to perform, in spite of how individuals perceive it. Moreover, considering this as an objective feature, it can also be more precisely measurable and increased, for it implies to materially act on characteristics that can be controlled from the outside. 

However, one of the main issues related to this concept is that it is still not very pervasive in the HRI research field. In fact, it is usually considered as overlapping with the figure of trust and this means that it is not still investigated by scientific community as much as it should.  

\section{A CHANGE IN PERSPECTIVE: BEYOND THE "NO TRUST-NO USE" PARADIGM}
The "no trust-no use" paradigm promotes an approach to social robotics that focuses solely on increasing confidence in machines, with the risk to underestimate the search for a balance with the side-effects of such a dynamic. 
In fact, it cannot be ignored that pathological consequences of HRI may include, among others: (i) over-trust, entrusting devices with tasks that go far beyond their actual functionalities \cite{sharkey2020we}; (ii) misrepresentation of the very nature of robots, inducing people to consider them as something more than mere objects; (iii) manipulation, interfering with the formation of human will \cite{palmerini2014robolaw}; (iv) impact on the user's risk-taken behaviour \cite{hanoch2021robot}. 

Moreover, although trust has various interpretations, all of them have in common the idea of this concept as a “multidimensional psychological attitude” \cite{jones1998experience}. Said otherwise, it is a properly human component. Despite all the scientific measures we can theorise or improve, there will always be a variable difficult to predict and even more difficult to control: human psyche.
Furthermore, trust represents a strictly relational propensity, that makes relevant the nature of the trustees themselves. In particular, they should have a positive disposition towards the trustor and should act freely and wittingly, so as to be held responsible in case of a breach of trust itself. In such a view, it could be accepted the claim of the part of the philosophical field which argues that none should trust robots, for they neither can feel genuine - human-like - empathy nor can be blamed for the negative effects of their behaviour. Nonetheless, it has been observed that people are actually inclined to have confidence in technological devices and that such a factor is a matter of fact in HRI. This is possible mostly because of the phenomenon of anthropomorphism which represents, as much as trust does, a natural and not-dismissible tendency, common to all human beings. That depends on the unconscious need to be relational and vulnerable to external influences. 

Therefore, it does not seem fruitful to argue about the desirability of trust in HRI, which in any case could not be removed. At the same time, focusing only on its enhancement and possible exploitation could collide with the priority to protect the physical and psychological integrity of people involved. Then, what we should do, for a responsible technological development, is to ponder \textit{how much} it is appropriate to emphasise such dynamics and to actually promote a "no trust-no use" approach to social robotics.
In order to solve this apparent impasse, this paper aims to suggest taking a step back in the investigation and change the perspective from which the theme of acceptability and development of social robots is addressed. 

That could be possible better delimiting the research question through which to conduct the analysis:  not whether we should or could trust machines, not even only how to do it more; yet to what extent and for which purpose it is suitable to trust robots. 

\section{DISCUSSION}
This paper provides insights to evaluate a rethinking of the role of trust in HRI, by asking to what extent and for which purposes it is fruitful to use trust to convey collaboration and usage.

Such a change in perspective could seem purely speculative. However, it would entail to start from a technically-specific analysis. Thus, the goal of any class of devices, their foreseeable and hypothetical effects on people’s psychological or material integrity should be considered. Then, it should be examined how to project them so as to hit their precise target, avoiding to \textit{over-design}.
It means to deal with the fact that people are able to trust machines and we cannot completely avoid this as a consequence of the interaction, because of dynamics that are inherent to every human being.

This is different from merely investigating how to make the user trusts the robot, or how to increase such a dynamic. 
In fact, it implies to focus on the identification of the \textit{minimum} level of trust, necessary for (i) an effective and efficient use of robotic devices and (ii) a mitigation - or even prevention - of the side-effects that can affect the users. This will allow to both favour technological development and guarantee that science will put human beings - as a whole - at the centre in such a development.

Moreover, if trust is a state of mind, mainly focusing on increasing it implicitly means to interfere with – and to manipulate – individuals' perception and the formation of individuals' will. This can expose them to non-negligible ethical issues. 
Therefore, it could be more worthwhile to focus on the concept of trustworthiness, for it is not a mere attitude, but a property of the object. Investing in material properties and quantitative analysis of social robots would imply making them more transparent, not just making them appear as such. 
This is a crucial aspect, considering that where transparency is incremented, the issues related with trust acquisition and trust maintenance in HRI could be more efficiently addressed \cite{lee2004trust}.
Actually, contrary to what the dominant research trend would suggest, trust and transparency are two alternative elements. Designing for transparency means designing for control \cite{botsman2017can}, instead of relying on a concept that is grounded more on personal and emotional elements than on a rational and controlled choice.
That does not mean to eliminate trust from the acceptability equation or to deny its relevance in robotics. It involves to suggest the possibility to rethink its role. 
Technical experts could focus on modulating the level of trust that guarantees the achievement of the settled goal for that technology, without undermining the protection of the user's integrity. 

Nevertheless, one of the main challenges in changing the perspective through which to evaluate the role of trust in social robotics could be the broad spectrum of disciplines that need to be taken into account. In fact, it would require an intensive multidisciplinary research. This implies to identify a delicate balance among very different fields, exponents of heterogeneous instances and methodological processes. 
In fact, it would need to structure an investigation which takes into rigorous account (i) technological factors - such as the peculiarities of different categories of devices, their respective purposes and contexts of application -; (ii) psychological and ethical implications - both those which are common to every human being and those which precisely regard the group targeted by the specific technology involved -.
Moreover, it could be fruitful to integrate future works with an analysis of the contribution that legal regulation may give to such a theme. In fact, on the 21th April 2021 the European Commission has proposed a draft of the first Artificial Intelligence Act (AIA) \cite{actproposal}, in which it appears clear that Europe favours a normative approach to the regulation of new technologies. 
In particular, it would be relevant to take into consideration the role of fundamental human rights. 
They have the advantage of being as flexible as ethical principles, but even as binding as legal norms. Moreover, they are proper of the humankind as a whole and inalienable. It follows that, with regards to the design and development of social robots, they could represent a balancing tool among the multidisciplinary instances of the disciplines involved, with a view to risk-benefit examination. 

Together with this, it could be useful to focus future researches on the concept of trustworthiness, so as to differentiate it more precisely from the concept of trust. This should be done in particular from a technical perspective, identifying material figures and scientific processes that can make social robots objectively trustworthy, instead of just subjectively trusted. 

\section{Acknowledgements}
The first author acknowledges that this work has received funding from the Alma Mater Research Institute for Human-Centered AI,  Law, Science and Technology Joint Doctorate, University of Bologna.
The second author acknowledges that this work is partially supported by the CHIST-ERA grant CHIST-ERA19-XAI-005, and by  FNR, the Luxembourg National Research Fund (G.A. INTER/CHIST/19/14589586).

\addtolength{\textheight}{-12cm}   % This command serves to balance the column lengths

%%%%%%%%%%%%%%%%%%%%%%%%%%%%%%%%%%%%%%%%%%%%%%%%%%%%%%%%%%%%%%%%%%%%%%%%%%%%%%%%

%%%%%%%%%%%%%%%%%%%%%%%%%%%%%%%%%%%%%%%%%%%%%%%%%%%%%%%%%%%%%%%%%%%%%%%%%%%%%%%%

%%%%%%%%%%%%%%%%%%%%%%%%%%%%%%%%%%%%%%%%%%%%%%%%%%%%%%%%%%%%%%%%%%%%%%%%%%%%%%%%

\bibliographystyle{plain}
\bibliography{Rachele}

\end{document}